%% file: main.tex
\documentclass{article}
\usepackage{spconf,amsmath,graphicx}
\usepackage{xspace}
\usepackage{bm}
\usepackage{paralist}
\usepackage{amssymb}
\usepackage{booktabs}

\newcommand{\method}{\textsc{LOGRAN}\xspace}
\ninept
\begin{document}
\title{Interpretable Multimodal Out-of-context Detection with Soft Logic Regularization}

%
\name{Huanhuan Ma$^{1,2, \ast}$,  Jinghao Zhang$^{1,2, \ast}$  Qiang Liu$^{1,2}$,  Shu Wu$^{1,2, \dagger}$, Liang Wang$^{1,2}$ \thanks{$^{\ast}$ The first two authors contributed equally.} \thanks{$^{\dagger}$ Corresponding author} }
  
\address{
    \textsuperscript{\textmd{1}} School of Artificial Intelligence, University of Chinese Academy of Sciences \\
    \textsuperscript{\textmd{2}} Center for Research on Intelligent Perception and Computing (CRIPAC)\\
    State Key Laboratory of Multimodal Artificial Intelligence Systems (MAIS)\\
    Institute of Automation, Chinese Academy of Sciences\\
    $\left\{huanhuan.ma, jinghao.zhang\right\}@cripac.ia.ac.cn, \left\{qiang.liu, shu.wu, wangliang\right\}@nlpr.ia.ac.cn$
    }

% \name{Daeun Kyung$^{1,\ast}$ \qquad Kyungmin Jo$^{1, \ast}$ \qquad Jaegul Choo$^1$ \qquad Joonseok Lee$^{2,3}$ \qquad Edward Choi$^{1, \dagger}$ \thanks{$^{\ast}$ The first two authors contributed equally.} \thanks{$^{\dagger}$ Corresponding author} \thanks{This work was supported by Institute of Information \& communications Technology Planning \& Evaluation (IITP) grant (No.2019-0-00075, No.2021-0-02068), Korea Health Industry Development Institute (KHIDI) grant (No.HI21C1138), Korea Medical Device Development Fund grant (Project Number: 1711138160, KMDF\_PR\_20200901\_0097), and National Research Foundation of Korea (NRF) grant (NRF-2021H1D3A2A03038607) funded by the Korea government(MSIT, MOTIE, MOHW, MFDS).}}
% \address{
%   $^1$KAIST \qquad $^2$Seoul National University \qquad $^3$Google Research
% }

%
\maketitle              % typeset the header of the contribution
\begin{abstract}

The rapid spread of information through mobile devices and media has led to the widespread of false or deceptive news, causing significant concerns in society. Among different types of misinformation, image repurposing, also known as out-of-context misinformation, remains highly prevalent and effective. However, current approaches for detecting out-of-context misinformation often lack interpretability and offer limited explanations.
In this study, we propose a logic regularization approach for out-of-context detection called \method (LOGic Regularization for out-of-context ANalysis).
The primary objective of \method is to decompose the out-of-context detection at the phrase level.
By employing latent variables for phrase-level predictions, the final prediction of the image-caption pair can be aggregated using logical rules. 
The latent variables also provide an explanation for how the final result is derived, making this fine-grained detection method inherently explanatory.
We evaluate the performance of \method on the NewsCLIPpings dataset, showcasing competitive overall results. Visualized examples also reveal faithful phrase-level predictions of out-of-context images, accompanied by explanations. This highlights the effectiveness of our approach in addressing out-of-context detection and enhancing interpretability.

\end{abstract}
\input{sec/intro}
\input{sec/relatedwork}
\input{sec/themethod}

\input{sec/exp}

\input{sec/conclusion}
\section{ACKNOWLEDGEMENTS}
This work is supported by National Natural Science Foundation of China (62372454, 62206291, 62236010).
% -------------------------------------------------------------------------

\bibliographystyle{IEEEbib}
\bibliography{mybibliography}

\end{document}

%% file: sec/intro.tex
\section{Introduction}

Due to the rise of mobile devices and the fast-paced development of media, information spreads more quickly than ever before. However, this rapid dissemination also enables the rapid spread of false or misleading news, causing significant concern about its negative consequences for society, individuals, and politics~\cite{nnycookBCR20,QianGSL18,XuLWW23,XuWLWW22,YuLWWT17,YuLWWT19,Wu0XW22,WuLLWT16}.
Some news make a false or misleading claim by using a real image to create more credible stories~\cite{sfooc,LuoDR21}. Specifically, the real images of people and events get reappropriated and used out-of-context to illustrate false events and misleading narratives by misrepresenting who is in the image, what is the context in which they appear, or where the event takes place. 
Image repurposing is unlike using deep fake models~\cite{RombachBLEO22,midjourney} to generate an image unexist, which potentially ampliﬁes its damage because of non-expertise need~\cite{ltms2020}. Such out-of-context misinformation can't be detected solely based on texts or images separately, thus automated detection methods are in high demand.

Existing research predominantly focuses on developing ``black box" models for predicting whether news articles involve the misuse of images in an out-of-context manner~\cite{sfooc,AbdelnabiHF22,lrnSourekAZK15}. However, these approaches often lack interpretability, providing only a holistic verdict without detailed explanations. 
Inspired by Chen et al.~\cite{chen2022loren}, we can detect a caption of a news image by its composing phrases, e.g., subject, verb, and object phrases. An image can be identified as out-of-context if it contains one or more unsupported phrases, which can be referred to as the culprits. Conversely, an image is deemed valid only when all the phrases in its caption are supported by the visual content it presents. 

It is also challenging due to the lack of phase-level labels in existing out-of-context datasets, such as the COSMOS dataset~\cite{sfooc} and the NewsCLIPpings dataset~\cite{LuoDR21}. This absence of labels makes it difficult for us to determine which specific aspect of the image caption renders it out-of-context. Manually annotating such ﬁne-grained data is unrealistic and requires tremendous human labor. 
The main concern is how to supervise a model to reach meaningful phrasal veracity predictions.
There are symbolic logical rules that can be utilized to create weak supervision for intermediate phrasal predictions. Empirically, all phrases of the image caption should not be detected as out-of-context use if the image is not out-of-context used, and a caption is detected as out-of-context use if there exists at least one out-of-context phrase. 

Motivated by the preceding description, we propose \method (LOGic Regularization for out-of-context ANalysis), a multimodal soft logic regularization approach designed to detect out-of-context instances for predicting the overall label, as well as to provide phrase-level predictions as explanations, thereby offering a comprehensive understanding of the detected instances.

\begin{figure}[t] %H为当前位置，!htb为忽略美学标准，htbp为浮动图形
\centering %图片居中
\includegraphics[width=0.48\textwidth]{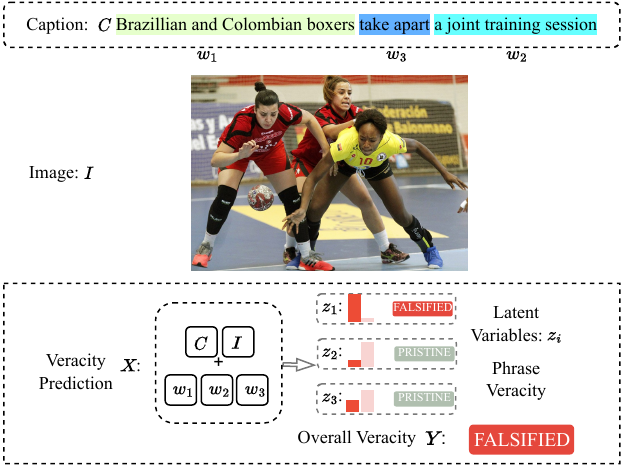} %插入图片，[]中设置图片大小，{}中是图片文件名
\caption{An example of the pipeline of how \method works. \method not only predicts the overall veracity of the image-caption pair, but also uses latent variables to indicate the veracity at phrase level. In this example, we can easily find the ``Culprit" of this caption is $w_1$. It manipulates the subject and the pristine caption is ``Tunisian and Angolan players fight for the ball on Sunday during a handball tournament in Spain Angola go on to win".}
\label{intro} %用于文内引用的标签
\end{figure}

Figure~\ref{intro} illustrates that the fundamental concept of \method is the use of a latent model. This method decomposes the given caption into a sequence of detectable phrases, and then employs a phrase-level prediction module as a latent model to generate the phrase predictions, treating these phrase predictions as latent variables. This approach is anchored in the framework of variational EM~\cite{qu2019probabilistic,ZhouHZLSXT20}, which contributes to generating meaningful predictions for the individual caption phrases.
Furthermore, \method incorporates logical rules into these latent variables by constructing a teacher module that aggregates logical reasoning across all latent variables. This process serves to distill logical knowledge into the target module. 

In the subsequent sections, we will conduct an in-depth exploration of our innovative approach, \method, followed by a comprehensive evaluation of its performance using various backbone architectures on the NewsCLIPpings dataset~\cite{LuoDR21}.

%% file: sec/relatedwork.tex
\section{Related Work}

\subsection{Multimodal Misinformation}

In response to the growing number of false claims associated with images, recent efforts have emerged to address this challenge and have introduced a few methods and datasets~\cite{McCraeWZ22,LuoDR21,sfooc}.

For instance, Luo et al.~\cite{LuoDR21} proposed a methodology for generating falsified examples by re-matching real images with real captions~\cite{vnews}. They created the extensive NewsCLIPpings dataset, which encompasses both authentic and convincingly falsified examples. This matching process was achieved automatically using the trained vision-language model CLIP~\cite{clip21}. 
The falsified examples generated could distort the context, location, or individuals depicted in the image, introducing inconsistencies in terms of entities and semantic context. The authors demonstrated that detection performances are limited, underscoring the challenging nature of the task. Similarly, Abdelnabi et al.~\cite{AbdelnabiHF22} proposed a method named CCN, which leverages external web evidence to verify image-caption claims.

In contrast, our approach takes a distinct direction towards interpretable out-of-context detection. We aim to identify where and how a caption has been falsified using a phrase-level approach. Our method does not rely on external content evidence for making predictions, and instead emphasizes explanations, thereby advancing the goal of being correct for the right reasons.

\subsection{Symbolic Logic Reasoning}

Previous researchers have attempted to combine symbolic logic and neural networks~\cite{lrnSourekAZK15,manhaeve2018deepproblog,lamb2020graph,wei2023multi}.
One prominent approach in this area relies on the variational EM framework~\cite{qu2019probabilistic,ZhouHZLSXT20}. Another commonly employed strategy involves incorporating neural network components to loosen the constraints of logic, enabling end-to-end training
% ~\cite{hu-etal-2016-harnessing,li-etal-2019-logic,wang2020integrating}.
~\cite{li-etal-2019-logic,wang2020integrating}.

Our approach follows a similar trajectory to Chen et al.~\cite{chen2022loren}, drawing inspiration from both of these approaches, where we combine elements from both research lines. Specifically, we utilize latent variables within a latent space to represent the intermediate predictions of truthfulness, mirroring the principles of the variational EM framework. These latent variables are subsequently subjected to regularization through a softened logic mechanism.

%% file: sec/themethod.tex
\section{The Method}
In this section, we introduce the proposed method \method for detecting whether an image-caption pair is out-of-context used. 
The possible prediction results are categorized as either \textit{Pristine} or \textit{Falsiﬁed}. Unlike most previous approaches that provide an overall prediction, our goal is to predict the final result as well as faithful phrase predictions for explanations. 
We define two tasks: overall caption detection and fine-grained phrase detection.

\textbf{Caption Detection}
Given a caption sentence $c$ and its image $I$, our goal is to model the probability distribution  $p(\bm{y}| c, I)$, where $\bm{y} \in \{\texttt{Pristine}, \texttt{Falsified} \}$ is a two-valued variable indicating the veracity of the caption's image.
We use $\bm{bold}$ letters to indicate variables in this paper.

\textbf{Phrase Detection}
We decompose the caption into phrases and predict the out-of-context label $\bm{z}_i$ for each caption phrase $w_i\in \mathcal{W}_c$ using the probability $p(\bm{z}_i|c, w_i, I)$, where $\bm{z}_i$ is treated as a binary latent variable $\bm{z}_i \in \{\texttt{Pristine}, \texttt{Falsified}\}$. 
We utilize the off-the-shelf tools provided by Flair and implement a phrase chunking model~\cite{chunking} to extract caption phrases.
The extracted caption phrases include named entities (NEs), verbs, adjectives, and noun phrases (APs and NPs).

\subsection{Overview of \method}

The \method approach offers more than just an overall prediction at the caption level; it also uncovers the veracity of individual phrases, providing insights into which part of the caption contributes to the final prediction.
We begin by following the variational EM framework~\cite{qu2019probabilistic,ZhouHZLSXT20} to build a latent model for phrase prediction. Each phrase prediction, denoted as $w_i \in \mathcal{W}_c$, is treated as a binary latent variable $\bm{z}_i$. We implement the negative Evidence Lower Bound (ELBO) loss $\mathcal{L}_\mathrm{var}(t, l)$ to optimize the latent model. At this stage, the latent variables can provide some degree of classification between \texttt{Pristine} and \texttt{Falsified} but still lack logical knowledge and constraints. Let $\bm{z} = (\bm{z}_1, \bm{z}_2, \dots, \bm{z}_{|\mathcal{W}_c|})$.  The final prediction of a caption $\bm{y}$ is then logically regularized based on the latent variables $\bm{z}$.
To integrate the logical knowledge into the latent variables, we propose a distillation method involving both a teacher model and a student model, along with the introduction of the logical loss $\mathcal{L}_\mathrm{logic}(t, l)$. The final loss function is a weighted sum of the $\mathcal{L}_\mathrm{var}(t, l)$ and the $\mathcal{L}_\mathrm{logic}(t, l)$.
In the following sections, we will provide a detailed introduction to our method.

\subsection{Latent Variable Model}
\label{sec:laten model}

Given an input $x = (c, E)$ comprising a textual caption $c$ and its corresponding image $E$, we define the target distribution $p_t(\bm{y}| x)$ as follows:
\begin{equation}
\label{eq:define}
    p_{t}(\bm{y} | x) = \sum_{\bm{z}} p_t(\bm{y} | \bm{z}, x) p(\bm{z} | x)
\end{equation}
Here, the target distribution $p_t(\bm{y}| x)$ as the sum of the probabilities of $\bm{y}$ given the latent variable $\bm{z}$ and $x$, multiplied by the prior distribution $p(\bm{z}|x)$. The independence assumption is made for $\bm{z}_i$, where $p(\bm{z}|x) = \prod_i p(\bm{z}_i | x, w_i)$.

The objective function is designed to minimize the negative log-likelihood of the gold label $y^\star$ given $x$. For optimization, we utilize variational inference and employ neural networks to amortize the variational posterior distribution. This is necessary due to the vast space of $\bm{z}$. Computing the exact posterior $p_t(\bm{z} | \bm{y}, x)$ poses a challenge for the EM algorithm.
This leads to minimizing the negative Evidence Lower BOund (ELBO), which consists of the expected log probability of $p_t(y^*|\bm{z}, x)$ under the variational posterior distribution $q_l(\bm{z} | \bm{y}, x)$ and the Kullback-Leibler divergence between $q_l(\bm{z}|\bm{y}, x)$ and $p(\bm{z} | x)$. The variational bound is denoted as $\mathcal{L}_\mathrm{var}(t, l)$:

\begin{equation}
\label{eq:elbo}
      -\mathbb{E}_{q_l}[\log  p_t(y^*|\bm{z}, x))] + D_\mathrm{KL}(q_l(\bm{z}|\bm{y}, x) \parallel p(\bm{z} | x))
\end{equation}

In our experiments, we utilize fine-tuned visual-language models as the prior distribution $p(\bm{z}|x)$. The features obtained from visual-language models are fed into a classifier, which provides some degree of classification between \texttt{Pristine} and \texttt{Falsified}.

\subsection{Distilling Logical Knowledge}
The ELBO (equation~\ref{eq:elbo}) alone does not ensure that the latent variables accurately represent the veracity of caption phrases without intermediate supervision. 
To address this issue, we need to integrate logical rules into the latent variables.
Here we declare the definition of the logical rules we used. Our insight is that certain natural logical consistencies need to be met between phrase detection and caption detection. 
Specifically, a caption is considered: 
\begin{inparaenum}[\it 1)]
    \item \texttt{Falsified} if there is inconsistency in at least one caption phrase;
    \item \texttt{Pristine} if all caption phrases are consistent. 
\end{inparaenum}

Having established the logical constraints, we propose a distillation method involving a teacher module and a student module to integrate the above logical rules into the latent variables. The student module, denoted as $p_t(\bm{y}|\bm{z}, x)$, is the module we aim to optimize. Meanwhile, the teacher module is constructed by projecting the variational distribution $q_l(\bm{z}|\bm{y}, x)$ into a subspace called $q_l^\mathrm{T}(\bm{y}_z|\bm{y}, x)$, which adheres to logical rules. This subspace, represented by $\bm{y}_z$, captures the logical aggregation of $\bm{z}$. By simulating the outputs of $q_l^\mathrm{T}$, we can transfer logical knowledge into $p_t$. The distillation loss is formulated as:
\begin{equation}
\label{eq:logicloss}
    \mathcal{L}_\mathrm{logic}(t, l) = D_\mathrm{KL}\left( p_t(\bm{y} | \bm{z}, x) \parallel q_l^\mathrm{T}(\bm{y}_z | \bm{y}, x) \right).
\end{equation}

The behavior of $q_l^\mathrm{T}$ during predictions reflects the information captured by the rule-regularized subspace, indicating the uncertain and probabilistic nature of the predictions \cite{chen-etal-2020-uncertain}. By minimizing the distillation loss $\mathcal{L}_\mathrm{logic}$ in Equation \ref{eq:logicloss}, the predictions of phrasal veracity are regularized by the aggregation logic, even without specific supervisions for caption phrases.

For the training process, we adopt soft logic \cite{li-etal-2019-logic} to relax the strict adherence to logic. We use product t-norms for differentiability during training and regularization of the latent variables. As discussed in Section \ref{sec:laten model}, given the probability of the caption phrase veracity $\bm{z}$, we logically aggregate them into $\bm{y}_z$ using the following equations (ignoring the input $x$ for simplicity):
\begin{equation}
\begin{split}
    q_l^\mathrm{T}(\bm{y}_z = \top) &= \prod_{i=1}^{|\bm{z}|} q_l(\bm{z}_i = \top), \\
    q_l^\mathrm{T}(\bm{y}_z = \bot) &= 1 - \prod_{i=1}^{|\bm{z}|} q_l(\bm{z}_i = \top)
\end{split}
\end{equation}

where $\sum_{\bm{y}_z} q_l^\mathrm{T}(\bm{y}_z) = 1$ and $\sum_{\bm{z}_i} q_l (\bm{z}_i) = 1$. Here, the symbols $\top$ and $\bot$ represent the labels \texttt{Pristine} and \texttt{Falsified}, respectively.

The final loss function is a weighted sum of two objectives, where $(1-\lambda)$ and $\lambda$ represent the relative importance of the variational loss and the logical loss, respectively:

\begin{equation}
\label{eq:loss}
    \mathcal{L}_\mathrm{final}(t, l) = (1 - \lambda )\mathcal{L}_\mathrm{var}(t, l) + \lambda \mathcal{L}_\mathrm{logic}(t, l),
\end{equation}

Here, $\lambda$ is a hyperparameter that determines the trade-off between the two objectives.

\subsection{Classifier}

We utilize neural networks to parameterize $p_t(\bm{y}|\bm{z}, x)$ and the variational distribution $q_l(\bm{z}|\bm{y}, x)$ for veracity prediction. These models are optimized using the variational EM algorithm and decoded iteratively. In this approach, a fine-tuned visual-language model serves as the representation decoder for both text and image inputs.

To generate joint representations for local hidden representations $\{\bm{h}^{(i)}_\mathrm{local} \in \mathbb{R}^d\}$ and global hidden representation $\bm{h}_\mathrm{global} \in \mathbb{R}^d$, we compute the Hadamard product ($\odot$) between each phrase $w_i \in W_c$ and the image $I$ for the local hidden representations. For the global hidden representation, we take the Hadamard product ($\odot$) between the overall caption $c$ and the image $I$.

After obtaining these representations, two-layer MLPs are employed to parameterize $p_t(\cdot)$ and $q_l(\cdot)$
\begin{equation}
\begin{split}
    \label{eq:model}
    p_t(\bm{y}|\bm{z}, x) &= \text{MLP}(\bm{h}_\mathrm{global} \odot \bm{h}^{(i)}_\mathrm{local}),\\
    q_l(\bm{z}|\bm{y}, x) &= \text{MLP}(\bm{y}  \odot \bm{h}^{(i)}_\mathrm{local})
\end{split}
\end{equation}
The label embeddings of $\bm{y}$ (ground truth $y^\star$ during training), $\bm{h}_\mathrm{phrase}^{(i)}$, and $\bm{h}_\mathrm{caption}$ are concatenated as input to $q_l(\bm{z}_i | \bm{y}, x)$, which outputs the probability of $\bm{z}_i$. Similarly, $(\bm{z}_1, \bm{z}_2, \dots, \bm{z}_\mathrm{\max})$ (padded to a maximum length) along with $\bm{h}_\mathrm{caption}$ and $\bm{h}_\mathrm{phrase}$ are concatenated as input to $p_t(\bm{y} | \bm{z}, x)$, which provides the distribution of $\bm{y}$. During training, both $q_l(\cdot)$ and $p_t(\cdot)$ are jointly optimized using the loss function described in Equation \ref{eq:loss}. 

During inference, we \emph{randomly} initialize $\bm{z}$ and iteratively decode $\bm{y}$ and $\bm{z}$ using $p_t (\bm{y}|\bm{z}, x)$ and $q_l (\bm{z}|\bm{y}, x)$ until convergence. As a result, we obtain both the final prediction $y$ and the latent variables $\bm{z}$, which serve as the phrasal explanation.

%% file: sec/exp.tex
\section{Experiments}

% \begin{table*}[h]
% \caption{Classification accuracy on the test set for the following models: (I)   VisualBERT-CC pretrained on the Conceptual Captions dataset, (II) VisualBERT-VN pretrained on the Visual News, (III) Multimodal CLIP, and (IV) CLIP with \method. }
% \vspace{-0.3cm}
% \label{tab:results_merged}
% \begin{center}
% % \begin{small}
% \resizebox*{\textwidth}{!}{
% \begin{tabular}{lcc|cc}
% \hline
%                               & VisualBERT-CC & VisualBERT-VN & CLIP  & CLIP-\method  \\ \hline
% (a) Semantics/CLIP T-I & 54.13         & 57.74         & 58.59 & 59.03 \\
% (b) Semantics/CLIP T-T  & 57.14         & 59.49         & 68.36 & 70.81 \\
% (c) Person/SBERT-WK T-T & 59.47     & 63.33         & 66.57 & 71.42 \\
% (d) Scene/ResNet Place        & 56.36         & 61.12         & 69.64 & 73.14 \\
% Merged/Balanced               & 54.82         & 58.63         & 67.27 & 70.51 \\ \hline
% \end{tabular}
% }
% % \end{small}
% \end{center}
% \vspace{-0.2cm}
% \end{table*}

\begin{table*}[h]
\caption{Classification accuracy on the test set for the following models: (I) VisualBERT, (II) VisualBERT with \method, (III) Multimodal CLIP, and (IV) CLIP with \method. The underlined portions represent improvements from \method}
\vspace{-0.3cm}
\label{tab:results_merged}
\begin{center}
\resizebox*{\textwidth}{!}{
\begin{tabular}{lcc|cc}
\hline
                              & VisualBERT & VisualBERT-\method & CLIP  & CLIP-\method  \\ \hline
(a) Semantics/CLIP Text-Image        & 55.12      &   \underline{56.88}                 & 58.59 & \underline{59.03} \\
(b) Semantics/CLIP Text-Text        & 53.47      &   \underline{55.62}                 & 68.36 & \underline{70.81} \\
(c) Person/SBERT-WK Text-Text       & 63.32      &   \underline{65.27}                 & 66.57 & \underline{71.42} \\
(d) Scene/ResNet Place        & 60.72      &   \underline{62.41}                 & 69.64 & \underline{73.14} \\
Merged/Balanced               & 61.32      &   \underline{63.18}                 & 67.27 & \underline{70.51} \\ \hline
\end{tabular}
}
\end{center}
\vspace{-0.2cm}
\end{table*}

\subsection{Dataset}

We use the NewsCLIPpings dataset~\cite{LuoDR21}, comprising both pristine and falsified images. It employs automation to match captions and images from the VisualNews~\cite{vnews} corpus, offering various subsets based on matching methods.

\subsection{Backbone models}
We use both CLIP~\cite{clip21} and VisualBERT\cite{visualbert} state-of-art models to evaluate the \method.
\begin{itemize}

    \item \textbf{CLIP}
     utilizes distinct encoders for processing images and text, which are trained to produce comparable representations for associated concepts. CLIP is pretrained on a large-scale dataset consisting of 400 million image-text pairs collected from the web. During training, a contrastive loss function is utilized to maximize the cosine similarity between true image-text pairs. This approach enables the model to learn meaningful associations between images and corresponding textual descriptions.
    
    \item \textbf{VisualBERT}
     is another multimodal model that integrates visual and textual information. It includes a sequence of Transformer layers that use self-attention to automatically align components of a given text input with specific regions in a corresponding image input. 
    Following the setting of NewsCLIPping~\cite{LuoDR21}, we use a Faster-RCNN model\cite{RenHG017} to extract image bounding box features.
    % Following the setting of NewsCLIPping~\cite{LuoDR21}, we employ two versions of VisualBERT model. One is pretrained on Conceptual Captions~\cite{captions} while another is pretrained on VisualNews~\cite{vnews}.
    
\end{itemize}

\subsection{Implementation Details}

We first fine-tune the two backbone models on the NewsCLIPpings dataset, then train our model \method.
We implemented our method in PyTorch. 
We ﬁne-tune both backbone models as we train the classiﬁers on the NewsCLIPpings dataset and optimize two models with the Adam~\cite{Kingma:2015us} optimizer, where the batch size is fixed at 64. When fine-tuning the backbone model, we use a learning rate of 5e-5 for the classifier and 5e-7 for other layers. 
% The optimal hyper-parameters were determined via grid search on the validation set. 
When training our model \method, we freeze the two backbone models and use a learning rate of 5e-7. 
% Besides, we stop training if accuracy on the validation set does not increase for 10 successive epochs. 

\subsection{Results and Discussion}

\begin{figure}[h] %H为当前位置，!htb为忽略美学标准，htbp为浮动图形
\centering %图片居中
\includegraphics[width=0.49\textwidth]{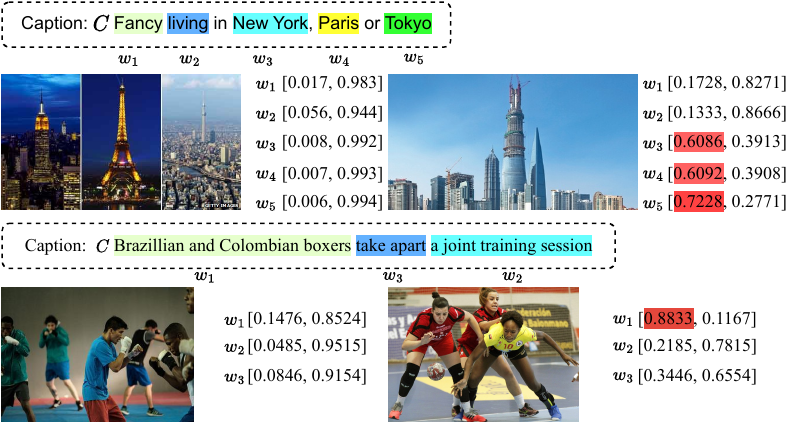} %插入图片，[]中设置图片大小，{}中是图片文件名
\caption{Some examples showcasing the outputs of \method. Each example includes a pair of images with matching captions. The left image is labeled as ``Pristine", while the right image is labeled as ``Falsified". We use the color red to highlight the phrases identified as ``Culprit".}
\label{result} %用于文内引用的标签
\end{figure}

Table~\ref{tab:results_merged} shows our experiments on the NewsCLIPpings dataset. The results indicate a significant improvement achieved by \method across all splits of NewsCLIPpings when compared to baseline methods without \method. 
Our proposed \method method demonstrates a significant improvement, achieving a three-percentage-point increase when utilizing the two backbone models. This finding demonstrates that our method can significantly improve overall classification accuracy while providing fine-grained explainability and robustness across various backbone models.

Additionally, we select specific cases to demonstrate the explainability of \method for phrase prediction, shown in figure~\ref{result}. The results also demonstrate that the phrase veracity predictions generated by \method exhibit interpretability, enabling us to identify the `Culprit' responsible for the falsified instance.

%% file: sec/conclusion.tex
\section{Conclusion}

This paper introduces \method, an interpretable Multimodal Out-of-context Detection method that not only makes overall predictions but also offers phrase predictions as explanations.
We achieve this by utilizing a variational EM framework to get latent variables and decomposing the out-of-context detection task at the phrase level. To incorporate logical rules into the latent variables, we propose a distillation method that involves a teacher model and a student model. 
This ensures that the latent variables accurately capture the veracity of caption phrases without relying on intermediate supervision.
To assess the efficacy of our approach, we have instantiated it in conjunction with two state-of-the-art visual-language models as the underlying backbone. We conducted extensive experiments on the NewsCLIPpings dataset, yielding compelling results and the robustness of our methodology.
Furthermore, we visualize some prediction results, which also demonstrate the interpretability of the \method.
Thereby emphasizing the contribution of our \method advancing the interpretability of out-of-context detection.